\documentclass{article}

\usepackage[final]{nips_2018}

\usepackage[utf8]{inputenc} 
\usepackage[T1]{fontenc}    
\usepackage{hyperref}       
\usepackage{url}            
\usepackage{booktabs}       
\usepackage{amsfonts}       
\usepackage{nicefrac}       
\usepackage{microtype}      
\usepackage{amsmath}
\usepackage{algorithm}
\usepackage{algorithmic}
\usepackage{amssymb}
\usepackage{graphicx}

\title{Bayesian Nonparametric Boolean Factor Models}

\author{
  Tammo Rukat\thanks{Work done while at the University of Oxford and the Alan Turing Institute} \\
  Amazon Research\\
  \texttt{tammorukat@gmail.com} \\
  \And
  Christopher Yau \\
  University of Birmingham and \\
  The Alan Turing Institute
}

\usepackage[dvipsnames]{xcolor}

\begin{document}

\maketitle

\begin{abstract}
  We build upon probabilistic models for Boolean Matrix and Boolean Tensor factorisation that have recently been shown to solve these problems with unprecedented accuracy and to enable posterior inference to scale to Billions of observations~\cite{Rukat2017, Rukat2018}. Here, we lift the restriction of a pre-specified number of latent dimensions by introducing an Indian Buffet Process prior over factor matrices. Not only does the full factor-conditional take a computationally convenient form due to the logical dependencies in the model, but also the posterior over the number of non-zero latent dimensions is remarkably simple. It amounts to counting the number false and true negative predictions, whereas positive predictions can be ignored. This constitutes a very transparent example of sampling-based posterior inference with an IBP prior and, importantly, lets us maintain extremely efficient inference. We discuss applications to simulated data, as well as to a real world data matrix with 6 Million entries.
\end{abstract}

\section{Introduction}

Boolean matrix factorisation decomposes binary data $\mathbf{X} = [x_{nd}] \in \{0,1\}^{N\times D}$ into a pair of low-rank, binary matrices $\mathbf{Z}=[z_{nl}] \in \{0,1\}^{N\times L}$ and $\mathbf{U} = [u_{dl}] \in \{0,1\}^{D\times L}$.
The data generating process is based on the Boolean product between binary matrices, which can be expressed through logical operations:
\begin{align}
  x_{nd} = \underset{l}{\texttt{OR}}\left[ \texttt{AND}(z_{nl}, u_{ld}) \right] \;. \label{eq:bool}
\end{align}
This model provides a framework for learning from binary data, similar to binary factor analysis or a clustering with joint assignments, where each observation is assigned to a subset of $L$ cluster centroids or codes.  
Here, one of the factor matrices represents a basis of binary codes, while the other contains indicator variables and provides a compact representation denoting the presence or absence of codes. A feature $x_{nd}$ takes a value of one if it equals one in any of the assigned codes. Note, that formally the designation of $\mathbf{U}$ and $\mathbf{Z}$ as codes or compact representation is arbitrary. They denote subsets of rows and subsets of columns, respectively, but their roles would simply interchange upon transposition of the data matrix. Recently, a probabilistic model for Boolean matrix factorisation has been introduced \cite{Rukat2017}, enabling sampling based posterior inference that scales to Billions of data points and outperforms previous approaches in finding accurate decompositions. In this workshop paper, we build upon this model and lift the restriction of a finite number of latent dimensions by specifying Indian Buffet Process (IBP) as prior over one of the factor matrices.

This approach has been studied in similar models \cite{wood2012_non-param,Meeds2007}. Nevertheless, our approach is methodologically interesting because the conditional distribution over the number of new latent dimensions takes an extremely simple, intuitive form as we show in Section~\ref{sec:limit}. Moreover, this work is of practical interest because it scales Bayesian nonparametric inference to very large datasets as we demonstrate on a moderately sized example of single-cell gene expression data with 6 Millions data-points in Section~\ref{sec:experiment}. We conclude the introduction with a brief description of the finite probabilistic model for Boolean Matrix Factorisation.

\subsection{Probabilistic Boolean Matrix Factorisation}

Denoting binary data as $\{0,1\}$ greatly simplifies notation in the following but is an otherwise arbitrary choice. We add i.i.d. Bernoulli noise at the observation level to the model in eq.~\eqref{eq:bool} to find a factorial likelihood of the form
\begin{align}
  p(x_{nd}|\mathbf{u},\mathbf{z},\lambda) = \sigma\left[\lambda \tilde{x}_{nd} \left(1-2\prod\limits_{l}\left(1-z_{nl}u_{ld}\right)\right) \right]\;. \label{eq:lik}
\end{align}
 The logistic sigmoid, \mbox{$\sigma(y) = 1/(1+\exp(-y))$}, leads to a convenient expression by virtue of its property, \mbox{$\sigma(-y) = 1-\sigma(y)$} togehter with the mapping from $\{0,1\}$ to $\{-1,1\}$, defined by \mbox{$\tilde{x} = 2x-1$}. The noise is controlled by a global parameter $\lambda \in \mathbb{R}^+$. Due to the deterministic logical dependencies among the variables, the full conditional distribution for any entry of the factor matrices
takes a simple form that lends itself to highly efficient computation:
\begin{equation}
    p(z_{nl}|.) = \sigma \bigg[\lambda \tilde{z}_{nl} \sum\limits_d \tilde{x}_{nd} u_{ld}\prod\limits_{l'\neq l} (1{-}z_{nl'}u_{l'd}){+}\text{logit}(p(z_{nl})) \bigg]\,. \label{eq:update_z}
\end{equation}
In particular, terms inside the sum over $d$, are known to be zero if any of the following two conditions holds:
(i) $u_{ld} = 0$; (ii) $\exists\, l' \neq l\text{, where } z_{nl} = u_{ld} = 1$.
Testing for these logical conditions can be implemented efficiently and in parallel. The noise parameter is updated after each sweep through the factors, setting it to its maximum likelihood estimate which is available in closed form, akin to a Monte Carlo EM algorithm~\cite{Rukat2017, Rukat2018}.

\section{Taking the Infinite Limit}\label{sec:limit}

We use the likelihood in eq.~\eqref{eq:lik} and specify an IBP prior on one of the factor matrices,
\begin{align}
  \mathbf{Z} \sim \text{IBP}(\alpha) \nonumber.
\end{align}

The IBP is a prior over binary matrices, where the entries in each column follow a Bernoulli distribution with parameter $\mu_l$ and where each $\mu_l$ is drawn independently from a Beta-distribution. It results from integrating out the $\mu_l$ and taking the limit as $L\rightarrow\infty$, such that the distribution has support over an infinite number of latent dimensions. For the purpose of this paper, we omit further details and refer the interested reader to Griffiths and Ghahramani~\cite{griffiths2011_indian-buffet-proces}. For the other factor, we use and independent Bernoulli prior,
\begin{align}
  p(U|q) = \prod\limits_{d,l} q^{u_{ld}}\,(1-q)^{1-u_{ld}}\;.
\end{align}
In order to retain a greater degree of symmetry between $U$ and $Z$, we could alternatively choose a finite Beta-Bernoulli prior over the independent columns of $U$. However, we refrain from doing so, because it would prohibit parallel inference for the rows of $U$ as described in~\cite{Rukat2017}.

The number of columns, $L$, is notionally infinite and, in practice, denotes the number of columns with at least a \textit{one}. The infinitely many other columns do not affect the likelihood and therefore do not need to be represented explicitly.
We define the number of ones per column as $ m_l = \sum_{n=1}^N z_{nl} $.
Similarly, $m_{-n,l}$ omits row $n$ in the summation, denoting the number of times feature $l$ has been applied to observations $n'\neq n$.
Next we describe the sampling procedure for $Z$, while samples from $U$ are drawn as previously described~\cite{Rukat2017}.

\subsection{Updates for existing codes}
  If $m_{-n,l} > 0$, we sample from the conditional as usual, but with the infinite Beta-Bernoulli prior, $p(z_{nl}{=}1|\mathbf{z}_{n,-l}) = \frac{m_{-n, l}}{N}$. In analogy to eq.~\eqref{eq:update_z}, we find
  \begin{align}
  p(z_{nl}=1|.) = \sigma\left[\text{logit}\left(\frac{m_{-n, l}}{N}\right) + \lambda \tilde{z}_{nl} \sum\limits_d \tilde{x}_{nd} u_{dl}
    \prod\limits_{l'\neq l} (1-z_{nl'} u_{dl'}) \right]\;. \label{eq:simple_updates}
  \end{align}
The prior contribution couples the rows of $Z$, such that updates can not be computed in parallel. 

\subsection{Sampling new codes}
In practice, we only need to represent columns with non-zero entries. However, we still need to sample from the remaining columns. Let $L'_n$ denote the number of columns of $Z$ that contain a $1$ only in row $n$ and change the notation such that let $L$ denotes the number of remaining columns with non-zero entries. We can compute the probability of $L'_n$ in order to sample the number of such columns. This corresponds to the number of new dishes ordered by customer $n$ and is independent of the other rows of $Z$ such that the conditional distribution is given by\begin{align}
\begin{aligned}
p(L'_n|.) &= p(L'_n|\mathbf{x}_n, \mathbf{z}_{n,l=1:L+L'_n}, \mathbf{U}_{d=1:D,\,l=1:L}) \\
          &\propto p(\mathbf{x}_n| \mathbf{z}_{n,l=1:L+L'_n}, \mathbf{U}_{d=1:D,\,l=1:L}, L'_n)\;
            p(L'_n)\;. 
            \label{eq:L_prior}
\end{aligned}
\end{align}
The prior is $\text{Poisson}(\frac{\alpha}{N})$, the likelihood factorises over $d$ and can be computed by marginalising over the new columns of $U$,

\begin{align}
\begin{aligned}
p(x_{nd}&|\mathbf{z}_{n,l=1:L+L'}, \mathbf{u}_{d,\,l=1:L}, L'_n) \\ &=
\sum\limits_{\mathbf{u}_{d,l=L+1:L'_n}}
  \sigma \left[
    \lambda \tilde{x}_{nd} \left(
    1-2\prod\limits_{l=1}^L (1-z_{nl}u_{ld}) \prod\limits_{l=L+1}^{L'_n} (1-u_{ld})
    \right) \right] \, p(\mathbf{u}_{d,l=L+1:L'_n})\;. \label{eq:marginalise_u}
\end{aligned}
\end{align}

Note, that for positive predictions, that is for $x_{nd}$, where $\exists\,l\le L'{:} z_{nl}u_{dl}{=}1$,
the term in parentheses is independent of the entries in the
new columns of $U$, i.e. in the product that runs from $l{=}L{+}1$ to $l{=}L'_n$. The intuition is, that the logical disjunction explaining these data-points
already emits a \textit{one}, independent of any additional arguments.
Taking the logarithm of the factorial likelihood, we have a sum over the two different types of matrix entries, $x_{nd}$: The positive predictions, $\bar{P}$, and the negative predictions, $\bar{N}$, defined as $x_{nd}$, where $\nexists\,l: z_{nl}u_{dl} = 1$. Terms for the positive predictions are independent of $L'_n$ and will cancel when normalising the probabilities for different values of $L'_n$. For the negative predictions we have two cases to consider: The Boolean disjunction emits a one, if any entry in the previously unused columns of $\mathbf{U}$ is one and emits a zero otherwise. There exists a single configuration for the latter case where all new entries are zero.
We thus have
\newcommand{\betanormOld}[1]{q^{-#1}}
\begin{equation}
\begin{aligned}
    &\sum\limits_{\bar{N}} \log p(\mathbf{x}_n| \mathbf{z}_{n,l=1:L+L'_n}, U_{d=1:D,l=1:L}, L'_n) \\ &=
    \sum\limits_{\bar{N}} \log \left[
        p(\mathbf{u}_{d,l=L+1:L'_n} = \mathbf{0})\,\sigma( - \lambda \tilde{x}_{nd}) +
        p(\mathbf{u}_{d,l=L+1:L'_n}\neq \mathbf{0})\,\sigma(\lambda \tilde{x}_{nd}) \right] \\ &=
    \sum\limits_{\bar{N}} \log \left[
        q^{L'_n}\,\sigma( - \lambda \tilde{x}_{nd}) +
        (1-q^{L'_n})\,\sigma(\lambda \tilde{x}_{nd}) \right] \\ &=
    \text{FN} \log\left[\betanormOld{L'_n} \sigma(-\lambda) +
                    \left(1{-}\betanormOld{L'_n}\right) \sigma(\lambda) \right]\, +
    \text{TN} \log\left[\betanormOld{L'_n} \sigma(\lambda) +
                    \left(1{-}\betanormOld{L'_n}\right) \sigma(-\lambda) \right]\;.
        \label{eq:negative_contributions}
\end{aligned}
\end{equation}

\begin{figure}
\begin{minipage}{.46\linewidth}
  \begin{algorithm}[H]
    \centering
    \caption{Sampling from $Z$ with IBP prior}
    \label{alg:ibp_alg}
    \begin{algorithmic}[tb]
      \FOR{$n=1,\ldots,N$}
        \FOR{$l=1,\ldots,L$}
          \IF{$m_{-n,l} > 0$}
            \STATE sample $z_{nl}$ from eq.~\eqref{eq:simple_updates}
          \ELSIF{$m_{-n,l} = 0$}
            \STATE Remove column $l$ from $Z$ and $U$.
          \ENDIF
        \ENDFOR
        \STATE Draw number of new $L'_n$ following eq.~\eqref{eq:L_prior}
        \STATE Set $z_{n,l=L+1:L'_n} = 1 $
        \FOR{$l=L+1,\ldots,L'_n$}
          \FOR{$d=1,\ldots,D$}
            \STATE sample $u_{dl}$ as previously (eq.~\eqref{eq:update_z})
          \ENDFOR
        \ENDFOR
      \ENDFOR
    \end{algorithmic}
  \end{algorithm}
\end{minipage}
\hspace{.04\linewidth}
\begin{minipage}{.5\linewidth}
    \centering
    \includegraphics[width=.8\linewidth]{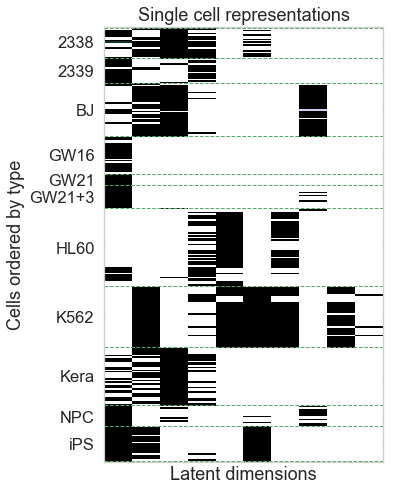}
    \caption{Binary representation (marginal posterior means) of 301 cell expression profiles (rows) ordered by cell type (black=1, white=0).
      \label{fig:sc}}
\end{minipage}%

\end{figure}

%
%
%

In the last step, we have subdivide the negative predictions into true negatives (TN), where $x_{nd}=0$ and false negatives (FN) where $x_{nd}=1$.
Note, that we can pre-compute the terms in the square brackets. These precomputed quantities need only be updated for a new values of $\lambda$. Thus, sampling $L'_n$ essentially amounts to counting the number of true positive and true negative predictions in the current configuration of the factors.
With the Poisson prior in eq.~\eqref{eq:L_prior} we can now compute the posterior probability for new values $L'$. We truncate the distribution over $L'$, by sampling only for $L'<10$.
The sampling procedure is sketched in Algorithm~\ref{alg:ibp_alg}.


\section{Experiments}\label{sec:experiment}

\subsection{Synthetic Data}
We generate synthetic data of size $200\times 500$ with balanced density from a Boolean product of iid random matrices, varying the latent dimensionality from 2 to 10. Figures for these experiments are not shown. 200 samples are drawn, the first 100 discarded as burn-in. We investigate posterios mean and modes of the distribution of latent dimensions, indicating the ability to recover the true data-generating dimensionality. We find that the model reliably recovers the ground-truth number of latent dimensions. In most cases, the sampler locks onto a single posterior mode.
We repeat these experiments adding noise with independent bit-flip probability to the data, where we find close-to perfect recovery for a noise level of 10\% and a systematic overestimation of roughly 1 latent dimension for a noise level of 20\%. The intuitive justification for this behavious is that the algorithms can not distinguish noise from \textit{true} patterns at this noise level and thus introduced additional latent patterns.

\subsection{Data from Single-Cell Gene Expression}
We show results for a real-world dataset from single-cell RNA expression analysis, a revolutionary experimental technique that facilitates the measurement of gene expression on the level of a single cell~\cite{Blainey2014}.
The dataset, described in~\cite{Pollen2014} consists of 301 cells of 9 known cell types. The number of sequencing reads per nucleotide is low such that we binarise data that now indicates the presence or absence of expression in approximately 21,000 genes with approximately 35\% of the cell/gene pairs being expressed. The data matrix has around 6 Million entries but drawing 200 samples from the factor matrices takes 1-2 minutes on a laptop. This is based on a Python implementation with substantial scope for further optimisation.
Figure \ref{fig:sc} shows the inferred cell-specific factor matrix. Each of the 301 rows depicts the marginal posterior mean of the binary representation a single cell profile. Each column is a latent dimension corresponds to a subset of the 21,000 genes. We see that the representation has a strong specificity for cell-types while some latent properties are shared across different cells. The corresponding gene-sets are biologically plausible.

\section{Conclusion and Future work}
We have shown that the probabilistic model for Boolean Matrix Factorisation~\cite{Rukat2017} can be efficiently extended using an IBP prior to infer a posterior distribution over the number of latent dimensions. Due to the logical structure of the posterior, computing full conditionals for Gibbs sampling is extremely fast. In particular, drawing samples from the distribution over additional latent dimensions amounts to counting the true negative and false negative predictions. The results is a flexible, nonparametric model for the analysis of binary data with outstanding scalability.

In future work we will extend this to data of arbitrary arity, as previously shown for the finite case~\cite{Rukat2018}. Moreover we will explore a fully parallel GPU-based implementation using the stick-breaking construction~\cite{Teh2007} as previously proposed~\cite{Zhang2017}.


\bibliographystyle{abbrv}

\begin{thebibliography}{1}

\bibitem{Blainey2014}
P.~C. Blainey and S.~R. Quake.
\newblock {Dissecting Genomic Diversity, One Cell At a Time.}
\newblock {\em Nat. Methods}, 11(1):19--21, jan 2014.

\bibitem{griffiths2011_indian-buffet-proces}
T.~L. Griffiths and Z.~Ghahramani.
\newblock {The Indian Buffet Process: An Introduction and Review}.
\newblock {\em J. Mach. Learn. Res.}, 12:1185--1224, jul 2011.

\bibitem{Meeds2007}
E.~Meeds, Z.~Ghahramani, R.~M. Neal, and S.~T. Roweis.
\newblock {Modeling dyadic data with binary latent factors}.
\newblock {\em Advances in neural information processing systems}, 19:977,
  2007.

\bibitem{Pollen2014}
A.~A. Pollen, T.~J. Nowakowski, J.~Shuga, X.~Wang, A.~A. Leyrat, J.~H. Lui,
  N.~Li, L.~Szpankowski, B.~Fowler, P.~Chen, et~al.
\newblock {Low-coverage single-cell mRNA sequencing reveals cellular
  heterogeneity and activated signaling pathways in developing cerebral
  cortex}.
\newblock {\em Nature biotechnology}, 32(10):1053, 2014.

\bibitem{Rukat2018}
T.~Rukat, C.~Holmes, and C.~Yau.
\newblock Probabilistic boolean tensor decomposition.
\newblock In J.~Dy and A.~Krause, editors, {\em Proceedings of the 35th
  International Conference on Machine Learning}, volume~80 of {\em Proceedings
  of Machine Learning Research}, pages 4413--4422, Stockholmsmässan, Stockholm
  Sweden, 10--15 Jul 2018. PMLR.

\bibitem{Rukat2017}
T.~Rukat, C.~C. Holmes, M.~K. Titsias, and C.~Yau.
\newblock {Bayesian Boolean Matrix Factorisation}.
\newblock {\em Proceedings of the 34th Annual International Conference on
  Machine Learning}, pages 2969--2978, jul 2017.

\bibitem{Teh2007}
Y.~W. Teh, D.~Gr{\"u}r, and Z.~Ghahramani.
\newblock Stick-breaking construction for the indian buffet process.
\newblock In M.~Meila and X.~Shen, editors, {\em Proceedings of the Eleventh
  International Conference on Artificial Intelligence and Statistics}, volume~2
  of {\em Proceedings of Machine Learning Research}, pages 556--563, San Juan,
  Puerto Rico, 21--24 Mar 2007. PMLR.

\bibitem{wood2012_non-param}
F.~Wood, T.~Griffiths, and Z.~Ghahramani.
\newblock {A Non-Parametric Bayesian Method for Inferring Hidden Causes}.
\newblock {\em arXiv preprint arXiv:1206.6865}, 2012.

\bibitem{Zhang2017}
Y.~L. Zhang, R.~C. Wang, K.~Cheng, B.~Z. Ring, and L.~Su.
\newblock Roles of {R}ap1 signaling in tumor cell migration and invasion.
\newblock {\em Cancer Biol Med}, 14(1):90--99, Feb 2017.

\end{thebibliography}

\end{document}